\documentclass[journal]{IEEEtran}
\usepackage{amsmath}
\usepackage{amsfonts}
\usepackage{amssymb}
\usepackage{bbm}
\usepackage[utf8x]{inputenc}
\usepackage{bm}
\usepackage{xspace}
\usepackage[font=footnotesize,figurename=Fig.,tablename=Tab.]{caption}
\usepackage[boxruled]{algorithm2e}
\usepackage{balance}
\usepackage{graphicx}
\usepackage{url}
\usepackage{siunitx}
\usepackage[hidelinks]{hyperref}
\usepackage{subcaption}
\usepackage{cite}
\graphicspath{{figures/}}
\usepackage{mathrsfs}
\usepackage{multirow}
\usepackage{textcomp}
\usepackage{etoolbox}

\newcommand{\SE}{SE(3)}
\usepackage{xcolor}

\title{\LARGE \bf
Achieving Dexterous Bidirectional Interaction in Uncertain Conditions for Medical Robotics 
}

\author{Carlo Tiseo, Quentin Rouxel, Martin Asenov, Keyhan~Kouhkiloui~Babarahmati,\\ Subramanian Ramamoorthy, Zhibin Li, and Michael Mistry 
\thanks{This work has been supported by EPSRC UK RAI Hub ORCA (EP/R026173/1), the Future AI and Robotics for Space (EP/R026092/1), National Centre for Nuclear Robotics (NCNR EPR02572X/1) and THING project in the EU Horizon 2020 (ICT-2017-1). For the purpose of open access, the author has applied a Creative Commons Attribution (CC
BY) license to any Author Accepted Manuscript version arising.}
\thanks{Carlo Tiseo, Quentin~Rouxel, Martin~Asenov, Keyhan~Kouhkiloui~Babarahmati, Subramanian Ramamoorthy, and Michael Mistry are with the Institute for Perception, Action, and Behaviour, School of Informatics, University of Edinburgh, UK. Zhibin Li is with the Department of Computer Science, University College London, UK. CT is also with the School of Engineering and Informatics, University of Sussex, UK. QR, 
Inria, CNRS, Universit\'e de Lorraine.{\tt\small c.tiseo@sussex.ac.uk}}
}

\begin{document}
\maketitle
\begin{abstract}
Medical robotics can help improve the reach of healthcare services. A  challenge for medical robots is their complex physical interaction. This work evaluates a recently introduced control architecture based on Fractal Impedance Control (FIC) in medical applications. The deployed FIC architecture is robust to delay between the master and the replica robots and can switch online between an admittance and impedance behaviour. Our experiments analyse three scenarios: teleoperated surgery, rehabilitation, and remote ultrasound scan. The experiments did not require any adjustment of the robot tuning, which is essential in medical applications where the operators do not have an engineering background. Our results show that it is possible to teleoperate the robot to perform remote occupational therapy, operate a scalpel, and use an ultrasound scan. However, our experiments also highlighted the need for a better robot embodiment to control the system precisely in 3D dynamic tasks. 
\end{abstract}

\section{Introduction}
Robot-mediated medical services have been identified as a possible solution to the ageing population in developed countries in the last few decades. An older population implies a lower active workforce and an increase in age-related diseases, increasing strain on the healthcare sector \cite{ReviewULR,ferraguti2015,ReviewRehab2,ReviewRehab3,feizi2021robotics,lambercy2021neurorehabilitation}. Additionally, as highlighted from the COVID-19 pandemic, reduced access to healthcare facilities can currently compromise healthcare quality. This problem was known in the sector, but it was not prioritised and was seen as a long-term problem because it affected only the population living in remote locations. The pandemic has revealed the short-term relevance of new technologies that can enhance the territorial permeability of these services.

Rehabilitation and robot-aided surgery are among the first applications in medical robotics\cite{hogan1992manus,ReviewULR}. The rehabilitation robots have shown how the introduction of these technologies in the rehabilitation centre has allowed an increase of bandwidth and therapeutic improvements in the patients \cite{hogan1992manus, ReviewULR, ReviewRehab2,ReviewRehab3}. Currently, multiple planar robots for upper-limb rehabilitation are available on the market, which can also be deployed at home or community centres \cite{lambercy2021neurorehabilitation}. Concurrently, the knowledge gained for rehabilitation robots supported the development of assistive technologies for medical, civil, and industrial applications. These technologies aim to support pathological cases, but they also target the reduction of injuries in the healthy population \cite{crea2021occupational}.

Surgical robots are the other devices that immediately attracted the attention of researchers, which have been seen as an opportunity to allow doctors to operate on patients remotely \cite{ferraguti2015,feizi2021robotics}. Endoscopic surgery also provided an ideal case study for robotics. Endoscopes were already an established device when roboticists approached the problem, providing minimally invasive access to internal organs, and they had established protocols and techniques \cite{tay2018use,ng2000development}. Therefore, medical robots could be developed to automate and improve an available technology, which has also increased the acceptability of these technologies in the medical community. An additional benefit of endoscopic surgery is the quasi-spherical operational field that can be projected on a flat screen without significantly impacting the operator's perception. More recently, the knowledge gained from developing co-bots, robots designed to share their workspace with humans, has also enabled the development of robots for orthopaedics surgery \cite{tian2020history}. In these systems, the doctor interacts with the end-effector to increase the quality of knee and hip prosthetics; however, the literature on these systems does not indicate a significant benefit of robot-aided surgery compared to traditional systems \cite{vanlommel2021initial}. 

Researchers have recently looked into performing other types of medical interventions in teleoperation, exploiting needles and scalpels\cite{ferraguti2015,feizi2021robotics}. Teleoperation presents unique challenges compared to autonomous manipulation.  The robot must follow the operator's real-time commands without knowing their intentions while maintaining interaction stability and adhering to safety constraints. The intrinsic interaction complexity connected with the variegated mechanical properties of biological tissues poses a challenge to traditional interaction control approaches, which rely on contact models. These controllers also require extensive tuning for switching between operations, requiring application-specific knowledge and profound knowledge of the control architecture. Furthermore, the introduction of delays and the exponential increase of computational complexity when multi-arms are involved render extremely challenging the applicability of these methods in teleoperation \cite{feizi2021robotics,minelli2019energy}. Therefore, these methods can potentially generate unsafe interaction due to the intrinsic energy tracking limitations due to the discrete nature of the virtual tank conservative energy \cite{babarahmati2019}.

\begin{figure*}[t]
      \centering
      \includegraphics[width=\textwidth, trim= 1cm 0.25cm 0.5cm 0.1cm,clip]{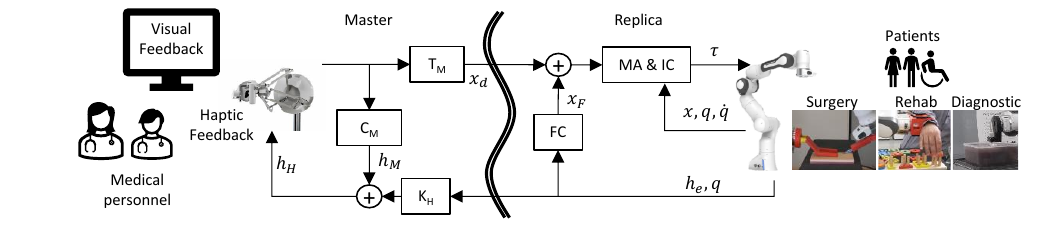}
      \caption{On the master side, there are the operator PC and the haptic feedback devices (Sigma.7, Force Dimension Inc.). On the Replica side, 7-dof torque-controlled arms (Panda, Franka Emika GmbH) are tested in scenarios targeting surgery, rehabilitation, and diagnostics. 
      The controller of the master has three elements. $\text{T}_\text{M}$ is the module that transforms the motion of the master ($\bm{x}_\text{M}$) in the desired pose for the replica ($\bm{x}_\text{d}$). $\text{C}_\text{M}$ is a controller providing virtual haptic feedback ($h_\text{M}$) to provide additional information to the user (e.g., workspace boundaries). $\text{K}_\text{H}\in [0,1] \subset \mathbb{R}$ is the gain applied to the wrench recorded at the end-effector of the replica robots ($h_\text{e}$). 
      The controller of the replica has two elements. $\text{FC}$ is the force controller that can be turned on when required, introducing an admittance controller on top of the low-level Interaction Controller (IC). $\text{MA \& IC}$ is a module composed of two components. The first element is the Motion Adaptation (MA) performed by an S-QP optimisation to guarantee that the desired trajectory respects the physical limitation of the robot (e.g., power limits and singularities) and the task (e.g., holding an object in bimanual manipulation). The second element is the IC that generates the torque commands to track the desired motion produced by the MA.
      It is worth remarking that in our experiments, the patients are substituted by two phantoms and a researcher, and another researcher acts as medical personnel.}
      \label{fig:Controller}
\end{figure*}
Other applications of co-bots in robotics involve automated diagnostics (e.g., ultrasound scan) and robot-aided TMS (Transcranial Magnetic Stimulation) \cite{pennimpede2013hot,noccaro2021development}. The automation of diagnostic technology looks into the possibility of completely automating examinations such as the ultrasound scan, looking into machine learning and neural networks to identify anomalies in the image and perform a diagnosis. The application for TMS aims to improve the stimulation by improving the neural circuit targeting, as this technology's effectiveness depends on the selective stimulation of the nervous tissues using magnetic induction.

Recently, our group has developed an impedance controller, called Fractal Impedance Controller (FIC), capable of robust interaction in unstructured environments without compromising the tracking accuracy \cite{babarahmati2019,Tiseo2021HapFic}. The FIC achieves these properties thanks to its passivity and path-independent observer, making it robust to delays and reducing bandwidth in state feedback \cite{babarahmati2020, tiseo2021fine,sena2021haptic}. The FIC teleoperation architecture has been experimentally tested in teleoperation for delays up to \SI{1}{\second} at a feedback bandwidth of \SI{10}{\hertz}, showing the robustness of interaction with a significant reduction of manipulability \cite{babarahmati2020}. The passivity also allows multiple controllers to be superimposed without affecting their stability, enabling decoupling the control problems and reducing the computational complexity\cite{Tiseo2021MIGame}. Earlier teleoperation experiments showcase how the proposed architecture enables the remote operator to collaborate with another person interacting with the replica robots\cite{babarahmati2020,tiseo2021fine,babarahmati2021robust,Wen2022}. 

This manuscript presents the preliminary evaluation of the performances of teleoperation architecture based on the FIC in using a scalpel, performing occupational therapy and an ultrasound scan (Fig.~\ref{fig:Controller}). The scope is to understand the potential capabilities of the proposed method and identify the challenges to overcome. Section \ref{Control} gives an overview of the controller, which is the same (including gains) presented in \cite{Wen2022}. Section \ref{Experiments} describes the experiments and presents the results. Sections \ref{Discussion} and \ref{Conclusion} discuss experimental results and draw conclusions, respectively.

\section{Control Overview} \label{Control}
The controller architecture comprises two sides with independent stability for their controllers, setting our control aside from other teleoperation architectures requiring their controllers' stability to be coupled \cite{babarahmati2021robust, Wen2022,ferraguti2015}. The master measures the motion of the user operator (e.g., medical personnel), using it as command input, and provides haptic and visual feedback from the replica side  (Fig. \ref{fig:Controller}). The replica reproduces the operator's movements and interacts with the patients and environment. This controller can operate one or multiple arms across various tasks by changing the end-effector mounted on the robots, as shown on the right side of Fig. \ref{fig:Controller}. It is worth noting that the arms can be either controlled independently or synchronised; notwithstanding the control modality, the stability of the two arms is independent, and their movements are synchronised, giving coordinated states for both effort and trajectory. Thus, we will present all the elements of the architecture for one robotic arm.

The master controller has three elements as described in Fig. \ref{fig:Controller}. $\text{T}_\text{M}$ generate the desired pose for the replica depending on the selected control mode. We implemented the position and velocity modes. The position mode passes the pose error of the master $\bm{x}_\text{M}\in \SE$ to the controllers of the replica device, reproducing it at the end-effector. It allows better dexterity in controlling the robot, limiting the workspace. The velocity mode updates the reference pose of the replica device via an integration of the velocity recorded at the master end-effector. The desired replica pose is the output $\bm{x}_\text{d}\in \SE$ defined as followed depending on the control mode:
\begin{equation}
\begin{array}{lc} 
    \bm{x}_\text{d}(t) = \bm{x}_\text{d}(0)+ \bm{x}_\text{M}(t),&\text{~~~ position mode}\\
    \bm{x}_\text{d}(t) = \bm{x}_{\text{d}} (t-1)+ \bm{\dot{x}}_\text{M}(t)\Delta t,&\text{~~~ velocity mode}
    \end{array}
\end{equation}
where $\bm{x}_\text{d}(0)\in \SE$ is the initial pose of the robot when the position mode is selected, $\dot{\bm{x}}_\text{M}\in \SE$ is the twist of the master device, and $\Delta t$ is the controller time step.
$\text{C}_\text{M}$ is the virtual haptic feedback ($h_\text{M} \in \mathbb{R}^6$) provided via the FIC-based controller NonLinear-PD (NLPD) formulated in \cite{Wen2022}, and it mimics the nonlinear controller acting on the replica that deals with unexpected interactions. It provides the haptic perception of the wrench (i.e., vector of force and torques) exerted on the robot end-effector by the user command. It also enhances the haptic information beyond the interaction force recorded on the replica, being able to provide feedback when the limits of the replica workspace are reached. This feedback is summed up as the wrench recorded at the end-effector ($\bm{h}_\text{e}\in \mathbb{R}^6$ ), scaled by a gain $K_\text{H}\in \left[0,~1\right]\subset \mathbb{R}$ that is controlled online by the user with the grasp-DoF of the Sigma-7 device. 

The replica controller has two main components, as shown in Fig. \ref{fig:Controller}. The Force Controller (FC) provides an admittance controller capable of tracking a desired interaction force at the end-effector. The Motion Adaptation (MA) and Interaction Controler (IC) adapt the desired motion ($x_\delta \in \mathbb{R}^6$) to the robot's physical capabilities and the task requirements and, subsequently, generate the torque command for the Replica ($\tau \in \mathbb{R}^{7}$).
The FC is an admittance controller that modifies the trajectory input by the user to account for the interaction at the robots' end-effectors, and it is based on the approach used in \cite{Wen2022}. It uses the the end-effector wrench ( $h_\text{e}$) and the joint kinematics ($q \in \mathbb{R}^7$) to estimate the displacement required to maintain the desired interaction force received by the MA.
The MA is performed with an algorithm called SEIKO Retargeting \cite{seiko,Wen2022}. This algorithm computes the desired whole-body configuration from the Cartesian input commands, solving a single iteration of Sequential Quadratic Programming (S-QP) at $1$ \si{\kilo\hertz} on the tangent space of the robot's trajectory. SEIKO Retargeting ensures that the next expected state is feasible (i.e., within the robots' kinematics and torque hardware limits) and does not pass through singularities. If any of these adverse conditions occur, the optimisation returns the feasible solution closest to the desired state.

The IC comprises a superimposition of five independent controllers, and all except the NLPD can be switched on and off without affecting the system's stability \cite{tiseo2023safe,Wen2022}. However, they might impact the accuracy and responsiveness of the replica. These controllers are a feed-forward load compensation, a postural joint space PD controller, a nonlinear compensation of the robot dynamics, and a relative Cartesian controller. This last controller is turned on only for the bimanual experiments, and it enhances the arms coordination.  

The multi-arm coordination can be switched on online, allowing the user to control multiple arms with a single haptic device. It is executed at the MA level, where the optimisation constraints are added to the conditions required to maintain the grip on the object. These constraints evaluate the contact forces with the object and the relative pose of the arms to maintain the grasp, derived by the grasp matrix and a simplified dynamic model of the object (i.e., geometry and mass)\cite{Wen2022}. Additionally, a fifth controller is turned on in the IC, which enforces the relative pose between the arms.  

\section{Experiments} \label{Experiments}
\begin{figure*}[t]
      \vspace{3mm}
      \centering
      \includegraphics[width=\textwidth, trim= 0cm 0cm 0cm 0cm,clip]{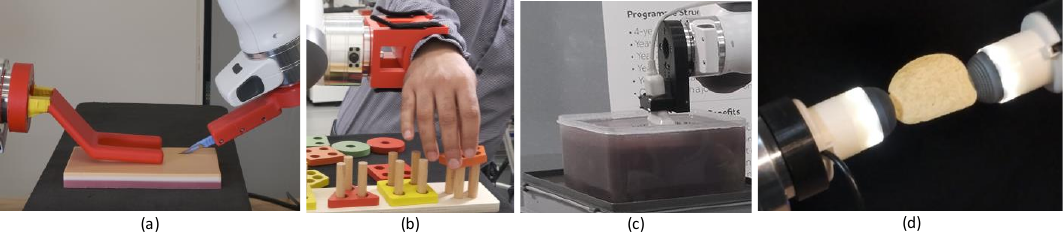}
      \caption{The proposed method has been used in multiple applications just by changing the end-effectors without requiring controller tuning. a) The hand end-effector used to hold the phantom during the cutting is mounted on the left arm and the support for the scalpel is on the right arm. b) The right arm has been equipped with a brace that is secure to the subject's arm with velcro straps. c) A vice-like end-effector is mounted on the right arm to secure the ultrasound probe to the robot. d) Two TACTIP sensors developed from the Bristol Robotics Laboratory \cite{pestell2019sense} have been mounted on the two robots to enable the bimanual telemanipulation of the potato chip.}
      \label{fig:End-Effectors}
\end{figure*}

We have designed four experiments to evaluate the capabilities of the proposed method in medical robotics and identify limitations. The first experiment targets surgery, and it is specific to using a scalpel to cut a silicone model of the human skin (Fig. \ref{fig:End-Effectors}a). The second experiment is a rehabilitation application, shown in Fig. \ref{fig:End-Effectors}b. It evaluates the system's capabilities to be deployed as a physical interface between the patient and the therapist during occupational therapy. The third experiment assesses the capabilities of the proposed method when performing an ultrasound scan (Fig. \ref{fig:End-Effectors}c). We used a phantom made of gelatin balloons (i.e., bladders, water and fruit. The fourth experiments evaluate the system's responsiveness in coordinated bi-manual manipulation of a fragile object (i.e., potato chip), evaluating the ability of the system to perform these tasks without reprogramming the controllers. However, it required the introduction of a soft force sensor at the end-effector (Fig. \ref{fig:End-Effectors}.d) to enhance the perception of the interaction forces. We also want to remark that we are focusing on the linear components of the control because the angular components are expressed in quaternion, and there is no intuitive way to visualise the results. 

\subsection{Scalpel Experiment}
The two end-effectors in Fig. \ref{fig:End-Effectors}a have been developed for this experiment. One end-effector holds the scalpel, and the other keeps in place the silicone skin during the cutting. The operator executes 16 cuts on the phantom, trying to proceed straight when crossing previously made incisions. The cross-incision is particularly interesting because a perpendicular cut weakens the phantom.  This experiment aims to test the dexterity of the system during cutting, evaluate the impact of the system manipulability on the task, and the visual and haptic feedback performances. 

The challenges of the scalpel experiment are the nonlinear soft-dynamics of interaction due both to the silicone of the phantom and the lateral flexibility of the scalpel blade, the 3D perception of the task before making contact, and the dexterity required to maintain the contact while executing long cuts in teleoperation.
 \begin{figure}[t]
      \centering
      \begin{subfigure}{1.0\linewidth}
		\centering
		\includegraphics[width=\columnwidth,, trim= 0cm 4.9cm 0cm 0cm,clip]{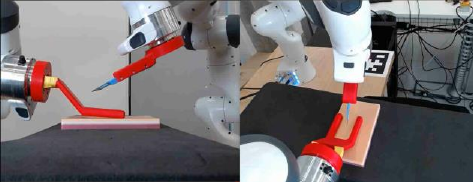}
	    \caption{Scalpel experiment}
        \label{fig:CameraViewsa}
	  \end{subfigure}\\
	   \begin{subfigure}{1.0\linewidth}
		\centering
		\includegraphics[width=\columnwidth,, trim= 0cm 1.25cm 0cm 3.65cm,clip]{Figure5.pdf}
		\caption{Rehabilitation experiment.}
        \label{fig:CameraViewsb}
	  \end{subfigure}
      \caption{Operator point of view for the scalpel and rehabilitation experiments.}
      \label{fig:CameraViews}
\end{figure}
\begin{figure}[t]
        \centering
        \begin{subfigure}{1.0\linewidth}
            \centering
            \includegraphics[scale=1, trim= 0cm 2.8cm .5cm 0cm,clip]{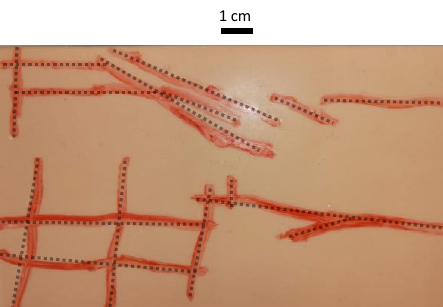}      
            \caption{}
            \label{fig:CutsOut}
	  \end{subfigure}\\
         \begin{subfigure}{1.0\linewidth}
		  \centering
            \includegraphics[scale=1, trim= 0cm 2.8cm 0cm 0cm,clip]{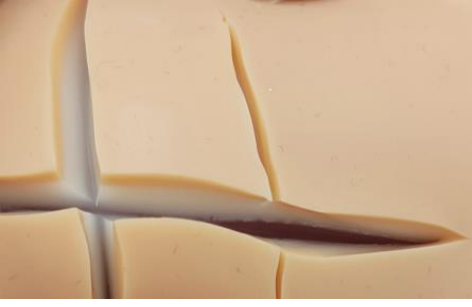}
            \caption{}
                \label{fig:CutsIn}
	  \end{subfigure}
        \caption{a) The cut marks on the silicone phantom show that it is difficult to proceed on a straight line. In addition, the deviation has peaks of a few millimetres, indicating the need to improve the system's performance on this task. b) The margins of the cut marks are needed, showing that the robot can robustly sustain contact with the phantom during the incision.}
      \label{fig:Cuts}
\end{figure}
This first experiment highlighted the difficulties in handling long-distance 3D movements with multi-camera views (Fig. \ref{fig:CameraViewsa}). Such an interface for the operator does not allow the concurrent perception of the movements in the two planes. Notwithstanding, the situation improves once the scalpel makes contact with the object and the task acquires a predominant planar component. Fig. \ref{fig:CutsOut} shows how the cuts have undulatory shapes around the segment connecting the start and the endpoints with deviations that can reach up to \SI{5}{\milli\meter} for longer cuts. However, the clean cuts on the material (Fig. \ref{fig:CutsIn}), the precise straight cut in some directions, and the presence of the undulatory behaviour in others seem to suggest that these deviations are due to the manipulability of the robots along this direction.

\begin{figure}[t]
      \centering
      \includegraphics[width=\columnwidth, trim= 6cm 10.5cm 6cm 10.2cm,clip]{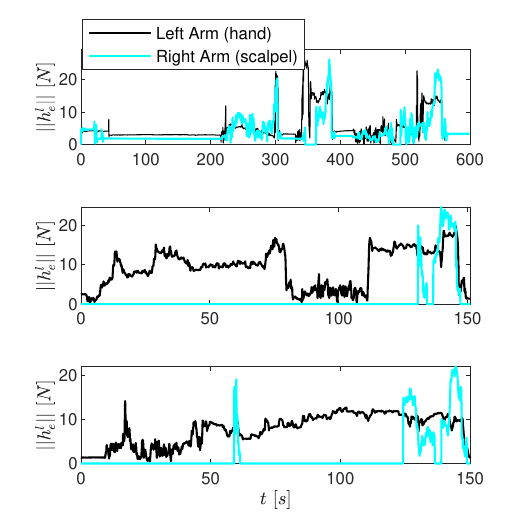}
      \caption{The force data for the scalpel experiments show that the robots are capable of sufficient force to hold the phantom down during cutting and can safely pass the peaks of force encountered during the cutting on the scalpel. The last two trials were conducted to check the performance in executing cross-cutting tests, and they were executed without changing the controller's parameters.}
      \label{fig:ForceScalpel}
\end{figure}

Finally, the operator also experienced difficulties with the haptic feedback for the left arm (i.e., hand end-effector), where there is the need for sustained interaction with the environment as shown in Fig. \ref{fig:ForceScalpel}. In contrast, this effect has a lower impact on the scalpel arm due to the reduced force peaks and shorter interactions' time, highlighted from the comparison of the signals' plot in Fig. \ref{fig:ForceScalpel}. Such a phenomenon can be explained by the low inertia of Sigma.7, which implies that the haptic feedback generates high tangential velocities in the master device. Thus, it requires active compensation from the user by increasing the interaction impedance and making the task tiring for the operator. 

\subsection{Rehabilitation Experiment}
The rehabilitation experiment is designed to evaluate the stability of the architecture when rigidly connected to a human via a brace while executing an activity simulating occupational therapy, as shown in Fig. \ref{fig:CameraViewsb}. 

The challenge of this test is the continuous trade-off between admittance and impedance behaviour. For example, if the patient takes the lead, the replica has to be transparent and behave as an admittance, while it has to switch to an impedance behaviour when the therapist intervenes to assist the patient. Such trade-off usually is extremely difficult for controllers because having an admittance controller acting on top of a non-rigid impedance controller tends to amplify the drift in the controller observer and eventually lead to instability. 

\begin{figure}[t]
      \centering
		\includegraphics[width=\columnwidth,trim= 0cm 3.5cm 3cm 2.4cm,clip]{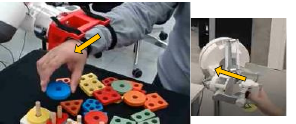}
       \caption{Snapshots of the master and the replica robots taken during assistance in the rehabilitation experiments.}
      \label{fig:RehabExperiment}
\end{figure}

\begin{figure}[t]
      \centering
      \includegraphics[width=\columnwidth, trim= 6cm 10.5cm 6cm 10.2cm,clip]{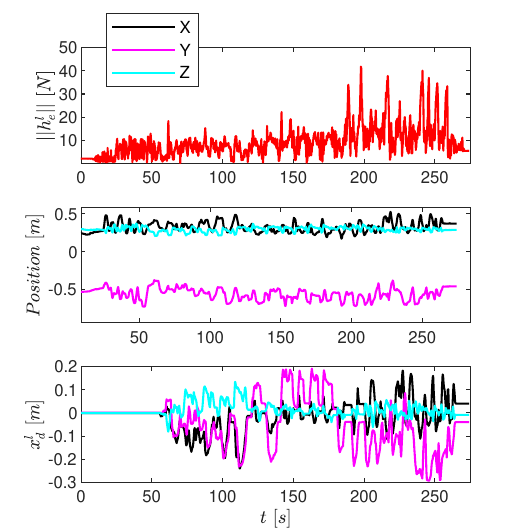}
      \caption{The norm of the force vector recorded in the first minute of the experiment shows that the robot can follow the patient movements with peak forces of about \SI{8}{\newton} once the \SI{2}{\newton} offset is accounted for. The forces recorded during assistance reach peaks of \SI{20}{\newton}, and they further increase close to \SI{40}{\newton} in the perturbation phase, which occurred for the last minute of the experiment.}
      \label{fig:RehabData}
\end{figure}

The experiment is divided into three tasks in Fig. \ref{fig:RehabExperiment}. The first investigates the transparency of the admittance controller to evaluate the level of compliance achievable without an additional end-effector force sensor by having the operator drive the robot to complete the task. The second task is about assistance with the operator assisting the patient in executing the task. Lastly, the third task is a disruptive interference from the operator, introducing perturbation to the user during the execution of the task.
The norm of interaction forces ($||\bm{h}_e^l||$) recorded in the experiment, the end-effector positions and the master controller linear command ($\bm{x}_d^l$) are shown in Fig. \ref{fig:RehabData}. The proposed method is capable of switching from the full admittance behaviour during independent movement where $||\bm{h}_e^l||$ peaks are about \SI{10}{\newton}. This occurs in the first minute of the experiment when $\bm{x}_d^l$ is close to 0. When the operator assists or perturbs the motions, the master error increases, generating a virtual force on the user. The assisted movements end when $t\approx\SI{180}{\second}$. It is characterised by higher interaction forces than autonomous motions, reaching peaks of about \SI{40}{\newton}. In contrast, in the perturbation phase, where there is an opposition to the subject's movements, the norm of the tangential force peaks.

\subsection{Ultrasound Scan Experiment}
The ultrasound scan experiment evaluates the ability of the system to perform a remote diagnostic test, which requires the design of an end-effector for holding the ultrasound probe (Fig. \ref{fig:End-Effectors}c). However, the available ultrasound scan does not allow remote control, so the experimental setup has been modified. This experiment is performed in the line of sight teleoperation. However, the phantom was placed above the operator's line of sight to hinder the perception and promote the use of the video feedback from the ultrasound scan. We use a phantom made of commercial food gelatin mixed with psyllium husk to enhance the contrast. We have three gelatin layers with different water components; the first has the recommended water-to-gelatin ratio. In the second layer, the amount of water is halved, and on the top layer, the water is reduced to one-third. Multiple props are suspended in the mix. There are bladders made of water balloons with grapes inside to mimic masses, and some fruit (grapes) is also distributed outside the bladder directly in the gelatine. It is worth noting that we have used a high-frequency probe that is not ideal for the quality of the image. However, it does not make any difference in evaluating the physical interaction stability and dexterity, which are the objectives of this experiment. 

\begin{figure}[t]
      \centering
      \includegraphics[width=\columnwidth, trim= 6cm 10.5cm 6cm 10.5cm,clip]{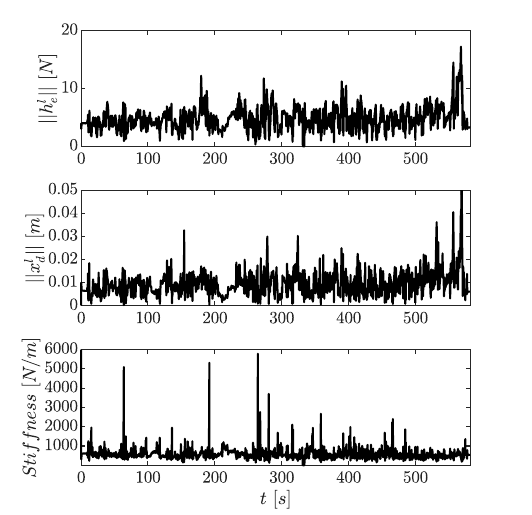}
      \caption{The forces, the tracking error and the end-effector stiffness of the replica robot during the ultrasound scan. It shows that the proposed method can adapt the impedance behaviour to adapt the changing non-linear dynamics at the end-effector.}
      \label{fig:DataUS}
\end{figure}

\begin{figure}[t]
      \centering
      \includegraphics[width=\columnwidth, trim= 0cm 4cm 0cm 0cm,clip]{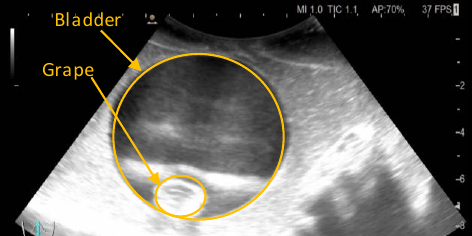}
      \caption{A screenshot of the ultrasound scan shows a water bladder with inside a grape.}
      \label{fig:USScan}
\end{figure}

Fig. \ref{fig:DataUS} shows the evolution of the interaction forces, the user input ($\bm{x}_d$) and the stiffness of the replica arm, showing how the proposed method can dynamically adjust its stiffness to interact with the nonlinear environmental dynamics. This autonomous modulation of the robot impedance allows stabilisation of the interaction while maintaining the required dexterity of interaction to perform the scan. The video also allows us to appreciate how, once the contact is made with the phantom, the exploration can be driven mainly relying on the ultrasound monitor shown in Fig.\ref{fig:USScan}. The main limitation of this experiment was that the available ultrasound did not allow remote adjusting of the image; thus, it required being physically close to the patient. 

\subsection{Bimanual Telemanipulation Experiment}
The bimanual teleoperation experiments are designed to test the responsiveness and the accuracy of the coordination during bimanual teleoperation (Fig. \ref{fig:End-Effectors}.d). We have chosen a potato chip because it is at the same time brittle, stiff, and light enough to make gravity a negligible component of the interaction forces. We introduce a soft end-effector capable of providing an indirect estimation of the interaction force. We have mounted a sensor developed by the Bristol Robotics Laboratory called TACTIP \cite{pestell2019sense}. It is based on a camera sensor placed inside a soft dome with a dotted geometric pattern, and the force estimation is obtained via the measurement of the pixel distance between the state of the deformed dome and the unperturbed state of the dotted geometrical pattern. The soft sensor detects subtle interaction forces, which could not be accurately estimated from the joint torques as in the previous experiments. The scope of this experiment is to test the system's transparent interaction performances by introducing an additional sensor at the robot end-effector.
\begin{figure}[t]
\vspace{5mm}
      \centering
      \begin{subfigure}{1.0\linewidth}
		\centering
		\includegraphics[width=\columnwidth,trim= 0cm 4.45cm 3.4cm 2cm,clip]{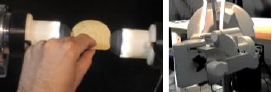}
		\caption{Object handover}
	  \end{subfigure}\\
	  \begin{subfigure}{1.0\linewidth}
		\centering
		\includegraphics[width=\columnwidth,trim= 0cm 2.4cm 3.4cm 4cm,clip]{Figure13.pdf}
		\caption{End-effector admittance driven interaction}
	  \end{subfigure}\\
	        \begin{subfigure}{1.0\linewidth}
		\centering
		\includegraphics[width=\columnwidth,trim= 0cm .4cm 3.4cm 6cm,clip]{Figure13.pdf}
		\caption{Teleoperated impedance driven bi-manual manipulation}
	  \end{subfigure}
      \caption{Snapshots from the Bi-manual telemanipulation experiments show the experiment's different phases. The end-effector mounted on the robot replaces the admittance controller, and the nonlinear dynamics of the dome substitutes the model-based state estimator.}
      \label{fig:PotatoChipSnapshot}
\end{figure}
\begin{figure}[t]
      \centering
      \includegraphics[width=\columnwidth, trim= 6cm 10.5cm 6cm 10cm,clip]{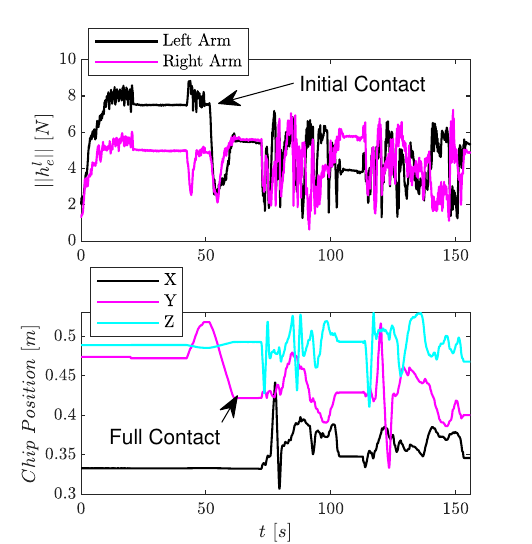}
      \caption{On top, the force at the end-effector is estimated from the joints' torques. On the bottom, the expected chip position before the contact and chip position after the contact. The plots highlight the need for the additional sensor at the end-effector. The differential interaction with the two arms is barely detectable and sometimes presents a bias in the equilibrium, as can be seen at $t=\SI{100}{\second}$.}
      \label{fig:PotatoChipData}
\end{figure}
Fig. \ref{fig:PotatoChipSnapshot} shows the different stages of the experiment, starting with the initial asymmetric contact made by the left arm, full contact once the right arm reached the potato chip, the user interaction with the object triggering the admittance behaviour of the controller, and the bi-manual telemanipulation showcasing the impedance behaviour of the proposed method. The interaction forces estimated from the joint torque compared with the position of the potato chip in Fig. \ref{fig:PotatoChipData} indicate that this is not a reliable way to estimate the interaction forces and the need to introduce the TACTIP end-effector for this task. We can observe multiple offsets in the contact forces estimated from the joints' torques (Fig. \ref{fig:PotatoChipData}) during and after contact with the objects. These offsets occur where the environmental interaction does not solely generate movements. This latter condition observable for about \SI{15}{\second} starting at $t\approx \SI{60}{\second}$ when the $||h_e^l||$ for the two arms are equal. Therefore, our experiments show that the flexibility of the proposed architecture allows overcoming the sensibility of the integrated admittance controller via the mounting of an instrumented end-effector that does not require any specific skillset in robotics. 

\section{Discussion} \label{Discussion}
The experimental results indicate that the proposed modular method is adaptable to multiple applications without tuning. The operator can change the target application by mounting the proper end-effector and selecting the associated architecture configuration. However, the modality selection is at the module level and does not require any tuning of the inner parameters of each module; thus, it is well suited for applications such as medical technologies where we have an expert operator with no engineering background.

The surgery experiments tested the possibility of establishing safe interaction with the soft tissues with a scalpel. However, the current limitations of the visual and haptic feedback need to be overcome before this technology can be comprehensively evaluated in experiments using more complex phantoms and biological samples. The deployment of virtual and augmented reality could help provide a better 3D perception, which will require studying the most-suited interface for providing comprehensive feedback and control of the system. Regarding the haptic feedback, employing the same robot in the master and the replica could help improve the user's feedback. This haptics is currently compromised by the high end-effector motions induced in the master device (Sigma.7) due to its lower end-effector inertia than the replica (Panda).

The rehabilitation experiment showcased how the controller's seamless trade-off between admittance and impedance behaviours allows robot-mediate collaboration between two human operators, which can also find application in other industries (e.g. manufacturing and logistics). Furthermore, it could also enable the deployment of commercial manipulators in rehabilitation, increasing the availability of robot-aided therapies and diversifying the market. Nevertheless, it also highlighted the same limitation in 3D perception in the visual feedback, which currently hinders assistance from the remote operator.

The ultrasound scan experiment showcased that it is possible to accurately control the probe for conducting a scan. The feedback from the scan monitor is sufficient to conduct the test once contact is made with the tissue; however, the 3D visual perception is essential to make contact with the desired anatomical district, which was possible thanks to the line of sight setup used for this experiment. The main limitation to the deployment of this technology is the lack of remote control for the ultrasound scan, which limits the distance of the operator from the patient to the length of the probe. Nevertheless, this application is currently the closest to eventual clinical testing among the evaluated scenarios. 

The bimanual telemanipulation experiments tested the possibility of having robot-mediated collaboration while manipulating fragile objects by introducing a sensorised end-effector to detect the low interaction forces at the end-effector. This application is still in early development, but the sensorised end-effectors could be used for assistive technologies and applications requiring the handling of delicate objects, such as in chemical laboratories. While all experiments presented in this work were conducted locally, \cite{sena2021haptic} demonstrated that our system can readily be applied to long-distance teleoperation over the internet, including multi-camera visual feedback with a latency of approximately \SI{200}{\milli\second}.

Lastly, another major limitation encountered in all the experiments is the limited embodiment of the remote arm, which makes it difficult for the operator to understand the manoeuvrability and the residual range of motion dictated both by the robot kinematics and the presence of objects. A possible option to enhance the embodiment is to exploit the virtual haptic controller in the master ($C_\text{m}$) to provide such information on the residual range of motion as a virtual resistive force.

\section{Conclusion} \label{Conclusion}
We presented a preliminary evaluation of a modular control architecture that enables the superimposition of manipulation and teleoperation in medical applications. Our experiments show that this method can provide robust physical interaction in a variegated set of scenarios without requiring a specialised robotic skill set to be reprogrammed. However, they also show perception issues in visual and haptic feedback, and they need to be improved before clinical testing. The visual feedback from a multi-camera view is not ideal for 3D dynamic tasks, which could be improved with an augmented reality interface. The haptic feedback is not ideal due to the gap of end-effector inertia between the master and the replica robots used in our experimental setup.
\balance
\bibliographystyle{IEEEtran}
\bibliography{main}
\end{document}